\documentclass[sigconf]{acmart}

\usepackage{multirow}
\usepackage{makecell}
\usepackage{subcaption}
\usepackage{xspace}
\usepackage{listings}
\usepackage{algorithm}
\usepackage{algorithmic}
\usepackage{placeins}
\usepackage{float}
\newcommand{\simraf}{\textbf{Sim-RAF}\xspace}
\newcommand{\miraf}{\textbf{MI-RAF}\xspace}

\usepackage{marvosym}

\AtBeginDocument{%
  }

\copyrightyear{2026}
\acmYear{2026}
\setcopyright{cc}
\setcctype{by}
\acmDOI{10.1145/3770855.3818897}
\acmConference[KDD '26]{Proceedings of the 32nd ACM SIGKDD Conference on Knowledge Discovery and Data Mining V.2}{August 09--13, 2026}{Jeju Island, Republic of Korea}
\acmBooktitle{Proceedings of the 32nd ACM SIGKDD Conference on Knowledge Discovery and Data Mining V.2 (KDD '26), August 09--13, 2026, Jeju Island, Republic of Korea}
\acmISBN{979-8-4007-2259-2/2026/08}

\begin{document}

\title{Retrieval-Augmented Foundation Models for Water Level Prediction in the Everglades}


\makeatletter
\def\@fnsymbol#1{\ensuremath{\ifcase#1\or *\or \#\or \dagger\or \ddagger\or
   \mathsection\or \mathparagraph\or \|\or **\or \dagger\dagger
   \or \ddagger\ddagger \else\@ctrerr\fi}}
\makeatother

\author{Rahuul Rangaraj}
\orcid{0009-0009-1897-646X}
\authornote{Co-first authors.}
\affiliation{%
  \institution{Florida International University}
  \city{Miami}
  \state{Florida}
  \country{USA}}
\email{rrang016@fiu.edu}

\author{Jimeng Shi}
\orcid{0009-0002-7268-0431}
\authornotemark[1]
\authornote{Co-corresponding authors.}
\affiliation{%
  \institution{Florida International University}
  \city{Miami}
  \state{Florida}
  \country{USA}}
\email{jshi008@fiu.edu} 

\author{Rajendra Paudel}
\orcid{0000-0002-7245-6985}
\affiliation{%
  \institution{Everglades National Park}
  \city{Homestead}
  \state{Florida}
  \country{USA}}
\email{rajendra\_paudel@nps.gov}

\author{Giri Narasimhan}
\orcid{0000-0003-0535-4871}
\authornotemark[2]
\affiliation{%
  \institution{Florida International University}
  \city{Miami}
  \state{Florida}
  \country{USA}}
\email{giri@fiu.edu}

\author{Yanzhao Wu}
\orcid{0000-0001-8761-5486}
\authornotemark[2]
\affiliation{%
  \institution{Florida International University}
  \city{Miami}
  \state{Florida}
  \country{USA}}
\email{yawu@fiu.edu}

%% ------------------------------------------------------------------
%% Abstract
%% ------------------------------------------------------------------
\begin{abstract}
Accurate water level forecasting in the Everglades is essential for flood mitigation, drought management, water resource planning, and biodiversity conservation. 
While recent time-series foundation models have shown strong performance on \textit{generic tasks} (represented in their pre-training), their effectiveness in \textit{domain-specific} applications remains insufficiently understood. 
In this work, we curate a domain-specific dataset for water-level forecasting in the Everglades and observe that the performance of current state-of-the-art models remains limited. 
To address this gap, we leverage a retrieval-augmented mechanism that retrieves analogous multivariate hydrological episodes from an external archive of historical observations to enrich the input context of those pre-trained models.
We study two retrieval strategies, statistical similarity-based retrieval and mutual information-based retrieval, and analyze how incorporating retrieved historical contexts affects predictive performance.
Extensive experiments show that retrieval augmentation consistently improves long-horizon water level forecasts and yields disproportionately larger gains during extreme events, which is particularly critical for environmental decision-making. 
Our study provides empirical evidence that analog-based retrieval can benefit pretrained time-series foundation models in environmental science, offering practical insights into their strengths, limitations, and failure modes when applied to hydrological forecasting in the Everglades. Although evaluated in the Everglades, the proposed framework is general and can be applied to other hydrological systems given time series data.
The code and data have been made publicly available at
{\url{https://github.com/rahuul2992000/WaterRAF}}.
\end{abstract}

\begin{CCSXML}
<ccs2012>
   <concept>
       <concept_id>10010405.10010432.10010437.10010438</concept_id>
       <concept_desc>Applied computing~Environmental sciences</concept_desc>
       <concept_significance>500</concept_significance>
       </concept>
   <concept>
       <concept_id>10002950.10003648.10003688.10003693</concept_id>
       <concept_desc>Mathematics of computing~Time series analysis</concept_desc>
       <concept_significance>300</concept_significance>
       </concept>
   <concept>
       <concept_id>10010147.10010257</concept_id>
       <concept_desc>Computing methodologies~Machine learning</concept_desc>
       <concept_significance>100</concept_significance>
       </concept>
 </ccs2012>
\end{CCSXML}

\ccsdesc[500]{Applied computing~Environmental sciences}
\ccsdesc[300]{Mathematics of computing~Time series analysis}
\ccsdesc[100]{Computing methodologies~Machine learning}

%% ------------------------------------------------------------------
%% Keywords
%% ------------------------------------------------------------------
\keywords{Water Level Prediction in the Everglades, Retrieval-Augmented Forecasting, Time Series Foundation Models}

%% ------------------------------------------------------------------
%% Make title
%% ------------------------------------------------------------------
\maketitle
%% ------------------------------------------------------------------
%% Main content
%% ------------------------------------------------------------------
\section{Introduction}
Everglades is a distinctive subtropical wetland ecosystem, spanning thousands of square miles in South Florida, recognized globally for its critical role in biodiversity conservation, regional flood mitigation, drought management, water resource planning, and sustaining local economies through recreation and infrastructure protection \cite{paudel2020assessing}. 
Given its ecological and socioeconomic importance, accurate water-level forecasting is critical for management, restoration, and operational decision-making in the region~\cite{saberski2022improved, wiederholt2020economic}.

Traditional hydrological models, including physics-based simulations \cite{sfwmd2005documentation} and statistical methods \cite{PEARLSTINE2020104783}, have long been used for water level prediction in the Everglades. However, physics-based models often require calibration of process parameters and can be computationally demanding. On the other hand, statistical models rely on specific assumptions and data conditions that limit their applicability. As a result, both approaches may not be suitable for transferring across changing hydrological regimes or extreme events. 
Recent advancements in deep learning (DL) have shown the potential to learn complex nonlinear interactions from data. For example, convolutional neural networks (CNNs), recurrent neural networks (RNNs), and Transformer-based models have achieved competitive performance across diverse time series forecasting tasks \cite{shi2025fidlar,kow2024advancing,shi2025deep,shi2025deep-weather-survey}. 
Despite the progress, these models need to be retrained whenever the experimental settings (e.g., different domains) are changed, resulting in domain-specific learning and limited generalization.

\begin{figure}[htbp]
\centering
    \includegraphics[width=0.47\textwidth]{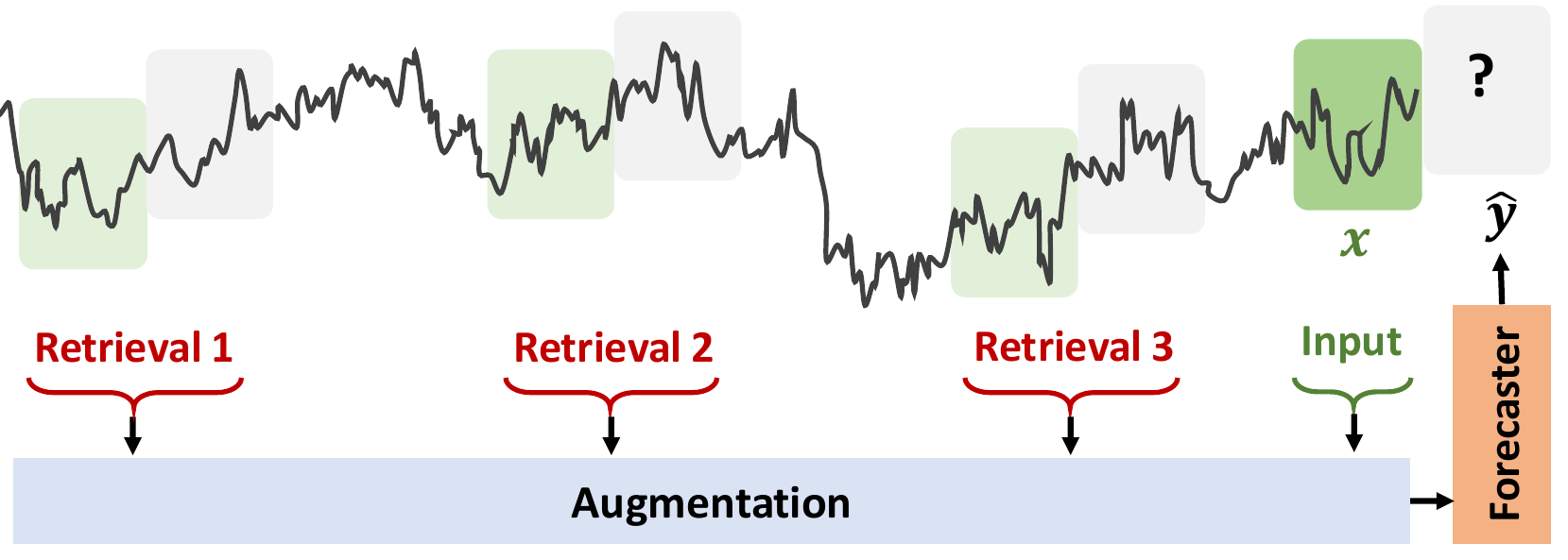}
    \caption{Illustration of retrieval-augmented forecasting (RAF). Traditional methods rely only on a recent lookback window $x$, whereas RAF augments $x$ with retrieved contexts.}
    \Description{A schematic diagram showing how retrieval-augmented forecasting adds retrieved past sequences to an input window before forecasting.}
    \label{fig:RAF}
\end{figure}

More recently, inspired by the transformative success of large language models (LLMs) in natural language processing, time series foundation models (TSFMs) \cite{miller2024survey,liang2024foundation} have emerged as powerful forecasting tools.
Similar to LLMs, TSFMs are pre-trained on massive and diverse time series datasets and can be adapted to a wide range of downstream tasks through zero-shot inference or lightweight fine-tuning.
Notable examples include TimeGPT \cite{garza2023timegpt}, TimesFM \cite{das2023decoder}, and Chronos \cite{ansari2024chronos}.
However, these models rely solely on information from a recent past window, which often contains limited and less informative signals, thereby restricting their predictive capability \cite{scussolini2024challenges}. 
It presents unique challenges for hydrological applications, such as water level forecasting in complex river systems, particularly in the face of rare or extreme events. 
For example, flood events may occur infrequently, providing little signal in immediate history for the model to learn from. Consequently, even the most advanced data-driven models are often ill-equipped to anticipate these low-frequency, high-impact events. 
To address this challenge, more informative input is needed.
Inspired by retrieval-augmented generation in NLP, \cite{lewis2020retrieval,fan2024survey} RAG methods augment the input with historically relevant contexts retrieved from an external archive.
Similarly, Retrieval Augmented Forecasting (RAF) methods \cite{han2025retrieval,zhang2025timeraf,yang2025timerag} have been explored in the time series field (as shown in Figure~\ref{fig:RAF}), which retrieve contextually relevant time series from a knowledge database and augment the original model input before forecasting.
By augmenting the model input with retrieved analogous scenarios, RAF enhances contextual awareness and predictive accuracy.
While RAF has shown promising results in various domains, including finance, energy, traffic, and weather \cite{tire2024retrieval,liu2024retrieval}, its application to time series forecasting in hydrology, in particular for the Everglades, remains largely underexplored, with limited analysis of long-horizon and extreme-event behavior.

In this work, we extended the Retrieval-Augmented Forecasting (RAF) paradigm to hydrological applications, tasked with water level forecasting in the Everglades. The objective is to predict future water levels based on an enriched input of the look-back window. Our approach consists of three key steps:
\textbf{(1) Retrieval:} Given the current input window, a retriever identifies high-relevance sequences from the knowledge base to serve as relevant contextual information.
Both similarity-based and mutual information-based retrieval approaches are explored due to their prevalence and strong performance \cite{cao2023mutual, zhang2025timeraf}.
\textbf{(2) Augmentation}: The retrieved sequences are integrated with the original input to form an augmented input representation.
\textbf{(3) Forecasting:} This augmented input is then passed to a time series foundation model to generate the water level forecasts. 
This approach allows forecasting models to access richer temporal context beyond the immediate past window, improving their ability to generalize under dynamic and extreme conditions, without requiring task-specific retraining.
This context-augmentation strategy is particularly relevant in non-stationary or data-sparse regimes, where statistical properties change over time, opening up a new direction for enhancing water level prediction in hydrologically complex and ecologically sensitive systems. Although our experiments focus on the Everglades, the proposed RAF pipeline is basin-agnostic and can transfer to other hydrological systems with comparable multivariate monitoring data and historical archives, while preserving the retrieval and evaluation protocol.
Our contributions are summarized as follows:
\begin{itemize}
\item \textbf{Dataset:} A domain-specific multivariate hydrological dataset from Everglades National Park, structured to support retrieval of historical analogs for forecasting.
\item \textbf{Methodology:} A retrieval-augmented forecasting framework that adapts pretrained time-series foundation models to hydrological forecasting by enriching inputs with historically relevant multivariate contexts, without task-specific retraining.
\item \textbf{Experimental Analysis:} A systematic comparison of statistical similarity-based and mutual information-based retrieval strategies, with detailed evaluation across stations, lead times, and extreme-event conditions. 
\item \textbf{Scientific Findings:} Empirical evidence that historical analogs substantially improve long-horizon forecasts and extreme-event prediction, revealing strengths and failure modes of foundation models under different hydrological conditions.
\end{itemize}
\section{Related Work}
\paragraph{Physics-based and Statistical Models.} Traditional hydrological models, including 1D/2D HEC-RAS (\cite{munoz2022inter}; \cite{huang2023alternative}) and region-specific systems like SFWMM \cite{sfwmd2005documentation}, have been pivotal in simulating water levels across various contexts. While such physics-based models remain foundational, they rely on calibrated parameters and static assumptions, limiting their adaptability during anomalous events. 
Statistical methods, such as Autoregressive Integrated Moving Average (ARIMA) \cite{agaj2024using} and seasonal ARIMA (SARIMA) \cite{azad2022water}
have proven useful for short-term forecasting and pattern identification. However, their ability to capture complex spatiotemporal dependencies is limited, particularly under extreme variability. 

\paragraph{Data-Driven Approaches.} 
Machine learning (ML) models have been largely used in hydrology for streamflow prediction, rainfall-runoff modeling, and groundwater level estimation \cite{khan2006application,nguyen2021development}.
Deep learning (DL) stands out by automatically learning relevant features from data. 
Recurrent Neural Networks \cite{shi2023explainable} and Convolutional Neural Networks \cite{bassah2025forecasting} have been applied to capture sequential and spatial dependencies in hydrological time series of water levels, precipitation, and river discharge \cite{yin2025physics,yin2023physic}.
The combination of Graph Neural Networks and Transformers \cite{shi2025fidlar} has been studied for flood prediction and mitigation in river systems.
The above methods are task-specific and need re-training when tasks and domains are changed.
More recently, the emergence of time series foundation models (TSFMs) marks a significant paradigm shift to general domains without re-training.
These models are pre-trained on massive, diverse time series datasets and can be adapted to downstream tasks with minimal data via zero-shot inference or light fine-tuning. 
Examples include TimeGPT \cite{garza2023timegpt}, TimesFM \cite{das2023decoder}, Chronos \cite{ansari2024chronos}, Moment \cite{goswami2024moment}, Moirai \cite{woo2024unified}, and Timer \cite{liu2024timer}, etc.
Recent efforts show the potential of TSFMs on hydrological forecasting \cite{rangaraj2025effective}.

\begin{figure*}[htbp]
\centering
    \includegraphics[width=0.98\textwidth]{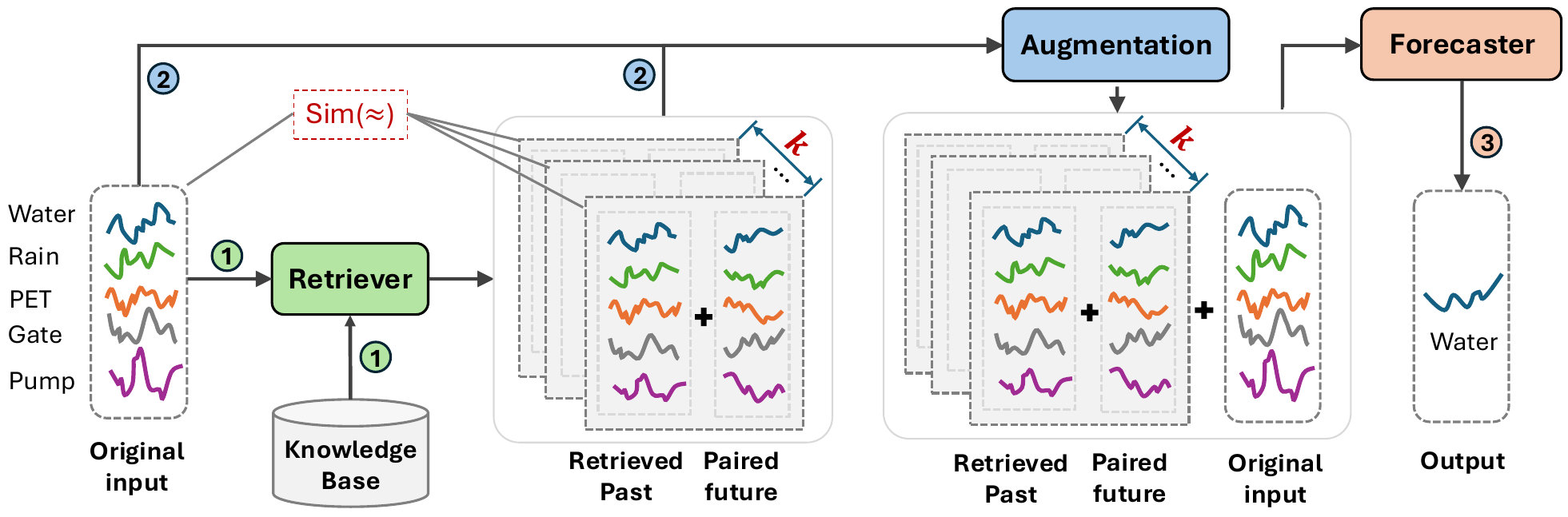}
\caption{Overview of Retrieval-Augmented Forecasting (RAF) framework for water level forecasting. 
Given a multivariate hydrological time series, \textbf{Retriever} fetches the top-$k$ similar sequences from a knowledge base, based on statistical affinity. 
The corresponding future observations are also retrieved alongside each context. 
\textbf{Augmentation} component enriches the original input by combining it with the top-$k$ context-future pairs. 
\textbf{Forecaster} takes the augmented input to make the final predictions. }
\Description{Diagram showing the three RAF components: retrieving top-k similar windows, augmenting input with retrieved contexts, and forecasting with a TSFM.}
\label{fig:framework}
\end{figure*}

\paragraph{Retrieval-Augmented Forecasting.} 
Originating in natural language processing, Retrieval-Augmented Generation (RAG) enhances foundation models by retrieving and incorporating external, contextually relevant data \cite{lewis2020retrieval}. 
Similar methods have been explored for time series forecasting, referred to as Retrieval-Augmented Forecasting (RAF). 
For example, RAF \cite{tire2024retrieval}, RATD \cite{liu2024retrieval}, RAFT \cite{han2025retrieval}, TimeRAF \cite{zhang2025timeraf}, and TimeRAG \cite{yang2025timerag} retrieve the relevant data by computing the similarity between the reference and candidate time series from the training set, demonstrating their effectiveness in boosting predictive accuracy.
However, the application of RAF to environmental forecasting, particularly in hydrology, remains largely unexplored.

\section{Methodology}
\label{sec:method}

We first introduce the water level forecasting problem. 
Let $\textbf{X}_{t-l+1:t} \in \mathbb{R}^{l \times m}$ denote a multivariate time series segment with $m$ hydrological variables (e.g., water levels, water flows, rainfall, Potential Evapotranspiration (PET)) observed over a lookback window of length $l$ at timestep $t$. 
Given the input $\textbf{X}$, the model $\mathcal{F}$ aims to predict the future sequence of water levels, ${\textbf{Y}}_{t+1:t+h} \in \mathbb{R}^{h \times n}$, for a horizon of length $h$ at $n$ locations of interest. 
Formally, it can be formulated as: 
\begin{equation}
    \mathcal{F}(\textbf{X}_{t-l+1:t}) \rightarrow {\textbf{Y}}_{t+1:t+h},
    \label{eq:water_forecast}
\end{equation}
where the subscripts represent the time range.

We then introduce key components in the proposed RAF framework for water level prediction, covering the selection of knowledge bases consisting of substantial samples (Sec.~\ref{sec:base}), 
retrievers devised to retrieve high-relevance samples from the knowledge base (Sec.~\ref{sec:retriever}), 
augmentation strategies to integrate the retrieved samples into the original input (Sec.~\ref{sec:augment}) 
and a time series foundation model as the forecaster (Sec.~\ref{sec:forecaster}). Figure~\ref{fig:framework} presents an overview.

\subsection{Knowledge Base}
\label{sec:base}
We briefly discuss the procedure for constructing a knowledge base, i.e., retrieval pool, for RAF. 
The preprocessing of all multivariate time series in the dataset is performed to construct samples comprising a pair of lookback window and a prediction window, denoted as $(\mathbf{X}_{\text{past}}, \mathbf{Y}_{\text{future}})$. 
The dataset for the water level prediction tasks is then split chronologically with a ratio of $85\%:15\%$. 
Within the RAF framework, the first 85\% serves as the knowledge base $\mathcal{D}$, from which the retriever selects high-relevance samples during the testing phase. The remaining 15\% of the samples are used for testing, $\mathcal{T}$.
They are represented as: 
\begin{align}
\mathcal{D} & = \left\{(\textbf{X}^{\text{base}(i)}_{\text{past}}, \textbf{Y}^{\text{base}(i)}_{\text{future}})\right\}_{i=1}^{d}, \label{eq:dataset} \\
\mathcal{T} & = \left\{(\textbf{X}^{\text{test}(j)}_{\text{past}}, \textbf{Y}^{\text{test}(j)}_{\text{future}})\right\}_{j=1}^{t}, \label{eq:testset}
\end{align}
where $d$ and $t$ represent the size of the knowledge base and test set, $i$ and $j$ are the indices of samples in knowledge base and test set.

\subsection{Knowledge Retrieval}
\label{sec:retriever}
Relying solely on the original lookback window $\textbf{X}^{(j)}_{\text{past}} \in \mathbb{R}^{l \times m}$ may be insufficient for accurately forecasting future water levels, ${\textbf{Y}}^{(j)}_{\text{future}} \in \mathbb{R}^{h \times n}$, as it provides limited contextual information. 
Therefore, a retriever is needed to identify a small number of analogous samples to enrich the input of a sole lookback window.
To this end, a scoring function is devised to evaluate the similarity or contextualization between the original lookback window $\textbf{X}^{(j)}_{\text{past}}$ and the candidate lookback windows $\textbf{X}^{(i)}_{\text{past}}$ from the knowledge base.
\begin{equation}
    \text{score} = S(\textbf{X}^{\text{base}(i)}_{\text{past}}, \textbf{X}^{\text{test}(j)}_{\text{past}}),
\end{equation}
where $S$ is the scoring function.
In the following, we present two different retrieval methods used in this work.

\paragraph{Similarity-based Retriever.}
It computes the similarity in the embedding space.
Following the Chronos work \cite{ansari2024chronos}, the T5 encoder \cite{rahman2020integrating} is employed to encode the lookback window of the original sample (i.e., original input) to an embedding vector with a fixed length, $k=512$.
The same encoding process is applied to all candidate lookback windows from the knowledge base.
Then we compute the Euclidean distance between these embedding vectors:
\begin{equation}
    \ell_2(\mathbf{e}^{\text{base}(i)}_{\text{past}}, \mathbf{e}^{\text{test}(j)}_{\text{past}}) = \left\| \mathbf{e}^{\text{base}(i)}_{\text{past}} - \mathbf{e}^{\text{test}(j)}_{\text{past}} \right\|%_2
\end{equation}

\paragraph{Mutual Information (MI)-based Retriever.}
MI, a fundamental concept from information theory that leans towards entropy, quantifies the statistical dependency between two random variables, capturing both linear and nonlinear relationships. 
In the context of hydrological forecasting, MI serves as a relevance signal to identify historical contexts most informative about the current system state.
The mutual information $MI(\textbf{X}^{\text{base}(i)}_{\text{past}}, \textbf{X}^{\text{test}(j)}_{\text{past}})$ is formally defined as:
\begin{equation}
MI = H(\textbf{X}^{\text{base}(i)}_{\text{past}}) + H(\textbf{X}^{\text{test}(j)}_{\text{past}}) - H(\textbf{X}^{\text{base}(i)}_{\text{past}}, \textbf{X}^{\text{test}(j)}_{\text{past}}),
\end{equation}
where $H$ denotes the entropy.

\paragraph{Ranking Retrieved Samples.}
All samples in the knowledge base are ranked based on corresponding scores (e.g., similarity or mutual information), and the top-$k$ highest-scoring samples are retrieved.
Each retrieved segment includes a lookback window and its corresponding future window. 
For clarity, we refer to the retrieved samples as a ``context'' set, 
\begin{equation}
    \mathcal{C} = \left\{(\textbf{X}^{\text{ctx}(z)}_{\text{past}}, \textbf{Y}^{\text{ctx}(z)}_{\text{future}})\right\}_{z=1}^{k},
    \label{eq:topk}
\end{equation}
where $k$ is the size of context set ($k \ll d$).

\subsection{Retrieval Augmentation}
\label{sec:augment}
\begin{figure}[ht!]
\centering
\includegraphics[width=0.98\columnwidth]{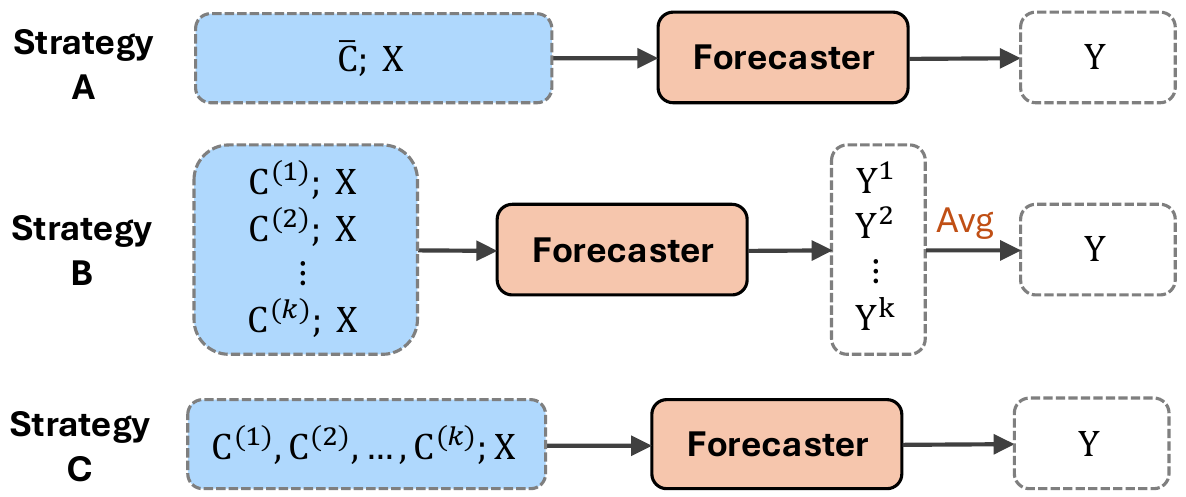}
\caption{Illustration of three augmentation strategies.}
\Description{Diagram showing Strategies A, B, and C for combining retrieved contexts with the original input sequence.}
\label{fig:stategies}
\end{figure}

To forecast water levels more accurately, the top-$k$ retrieved samples are integrated with the original input to form an augmented input. We explore three augmentation strategies as Figure~\ref{fig:stategies} shows.

\paragraph{Strategy A: Averaging-First, Augmenting-Later.}
In this approach, we compute the pointwise average over the top-$k$ retrieved candidates, $\mathcal{C}$, as shown in Eq. (\ref{eq:topk}) by:
\begin{equation}
    \bar{\mathbf{C}} = \frac{1}{k} \sum_{z=1}^{k} \mathcal{C}^{(z)}.
    % \bar{\mathbf{X}}_h = \frac{1}{k} \sum_{z=1}^{k} (\mathbf{X}_{\text{past}}^{\text{ctx}(z)}, \mathbf{Y}_{\text{future}}^{\text{ctx}(z)})
    \label{eq:average_topk}
\end{equation}
Then the averaged context is concatenated with the original input, $\mathbf{X}_{t-l+1:t}$, to form the augmented input: $\mathbf{X}_{\text{aug}}=[\bar{\mathbf{C}}; \mathbf{X}_{t-l+1:t}]$, where ``$;$'' denotes temporal concatenation.

\paragraph{Strategy B: Augmenting-First, Averaging-Later.}
Each candidate in the context set, $\mathcal{C}^{(z)}$, is prepended to the current input individually to form the $k$ augmented inputs $\mathbf{X}^{(z)}_{\text{aug}}=[\mathcal{C}^{(z)}; \mathbf{X}_{t-l+1:t}]$, where $z \in [1, 2, \dots, k]$.
The forecaster processes each augmented input individually, generating $k$ predictions that are subsequently averaged for the final forecast.

\paragraph{Strategy C: Long-Context Concatenation.} In this approach, all $k$ retrieved contexts are directly concatenated with the current input to form a single augmented input:
\begin{equation}
    \mathbf{X}_{\text{aug}} = [\mathbf{C}^{(1)}; \mathbf{C}^{(2)}; \dots ; \mathbf{C}^{(k)}; \mathbf{X}_{t-l+1:t}].
\end{equation}

\subsection{Water Level Forecaster}
\label{sec:forecaster}
In this study, we adopt Chronos \cite{ansari2024chronos} as the forecasting model due to its strong empirical performance in prior comparative studies on water level prediction tasks \cite{rangaraj2025effective}. Nevertheless, our RAF framework is model-agnostic and can be integrated with other pretrained or fine-tuned time series forecasting models. 
Built on the strong time series foundation model, Chronos, we introduce similarity-based and mutual information-based RAF methods, referred to as \textbf{Sim-RAF} and \textbf{MI-RAF} for brevity.

\begin{table*}[ht]
    \centering
    \caption{Summary of the dataset.}
    \resizebox{0.92\textwidth}{!}{ % Resize table to fit page width
    \begin{tabular}{l|c|c|c|l}
        \toprule
        \textbf{Feature} & \textbf{Interval} & \textbf{Unit} & \textbf{\#Var.} & \textbf{Location} \\
        \midrule
        Rainfall & Daily & $in$ & 2 & NP-205, P33 \\
        % \midrule
        Potential Evapotranspiration & Daily & $mm$ & 2 & NP-205, P33 \\
        % \midrule
        Pump flows & Daily & $cfs$ & 8 & S356, S332B, S332BN, S332C, S332DX1, S332D, S200, S199 \\
        % \midrule
        Gate flows & Daily & $cfs$ & 11 & S343A, S343B, S12A, S12B, S12C, S12D, S333N, S333, S334, S344, S18C \\
        % \midrule
        Water levels & Daily & $ft$ & 14 & S12A, S12B, S12C, S12D, S333, S334, NESRS1, NESRS2, NP-205, P33, G620, SWEVER4, TSH, NP62 \\
        \bottomrule
    \end{tabular}
    }
    
    \label{tab:data_summary}
\end{table*}

\section{Experimental Setup}
\subsection{Study Area and Dataset}
\paragraph{Study Area.} The study area focuses on the Everglades National Park, a distinctive subtropical wetland ecosystem vital for ecological balance and socio-economic activities in South Florida (refer to the detailed area map provided in Appendix \ref{sec:study_area}). 
Within this landscape, we monitored water levels at five key measuring stations, strategically located near major canals and water control structures, and within natural wetlands. The data in these stations are collected by Everglades National Park and are available in the Everglades
Depth Estimation Network (EDEN)~\cite{haider2020everglades} and the South Florida Water Management District’s Environmental Database (DBHYDRO)~\cite{sfwmd2024}, providing  daily measurements that underpin our forecasting models. 

\paragraph{Dataset.} 
Our dataset spans daily records from October 16, 2020, to December 31, 2024, comprising 1538 days and 38 hydrological and meteorological variables. It includes heterogeneous variables for a comprehensive study, such as water levels across multiple operational monitoring stations, rainfall, Potential Evapotranspiration (PET), and flow across hydraulic structures (gates and pumps). The main reason for choosing this period was to maintain consistency with water management operations. This period reflects the Combined Operational Plan, implemented in 2020 to deliver water to Everglades National Park, and its operations differ from those under previous plans. 
Data was extracted from the Everglades Depth Estimation Network (EDEN) \cite{haider2020everglades} and the South Florida Water Management District’s Environmental Database (DBHYDRO) \cite{sfwmd2024}. 
Table~\ref{tab:data_summary} summarizes the variables, units, temporal resolution, and spatial coverage of the dataset.
The data was split into training, validation, and test sets with an approximate ratio of $70\%:15\%:15\%$.
Pretrained foundation models and our RAF methods directly use the full 85\% knowledge base without a validation split. 
All models are evaluated on the same final 15\%, ensuring a consistent and fair comparison. The training set includes data from October 16, 2020, to May 14, 2024, representing 1,307 days, and the test set includes data from May 15, 2024, to December 31, 2024, comprising 231 days.

\begin{table*}[ht!]
\centering
\caption{Performance across 5 water stations for various lead times. The first four baselines are task-specific, followed by four pre-trained time series foundation models.
    The last two are our RAF methods with similarity-based and mutual information-based implementations utilizing the \textit{}{Strategy B} discussed in Sec. \ref{sec:augment} for retrieval augmentation. The best results are in \textbf{bold}. Additional RAF results comparing Strategy A and Strategy C are reported in Table~\ref{tab:stratA_C_results} in the Appendix.}
    \resizebox{1.75\columnwidth}{!}{
    \begin{tabular}{c|c|cc|cc|cc|cc|cc|cc}
        \toprule
        \multirow{2}{*}{Models} & \multirow{2}{*}{\makecell[c]{Lead\\Time}} 
        & \multicolumn{2}{c|}{NP205} & \multicolumn{2}{c|}{P33} 
        & \multicolumn{2}{c|}{G620} & \multicolumn{2}{c|}{NESRS1} 
        & \multicolumn{2}{c|}{NESRS2} & \multicolumn{2}{c}{Overall} \\
        \cmidrule(lr){3-4} \cmidrule(lr){5-6} \cmidrule(lr){7-8} 
        \cmidrule(lr){9-10} \cmidrule(lr){11-12} \cmidrule(lr){13-14}
        & & MAE & RMSE & MAE & RMSE & MAE & RMSE & MAE & RMSE & MAE & RMSE & MAE & RMSE \\

        \midrule
        \multirow{4}{*}{\makecell[c]{DLinear\\ \cite{zeng2023transformers}}} & 7  & 0.140 & 0.235 & 0.087 & 0.129 & 0.097 & 0.148 & 0.101 & 0.145 & 0.101 & 0.142 & 0.105 & 0.160 \\
                & 14 & 0.203 & 0.320 & 0.123 & 0.168 & 0.144 & 0.199 & 0.144 & 0.192 & 0.146 & 0.191 & 0.152 & 0.214 \\
                & 21 & 0.305 & 0.416 & 0.246 & 0.282 & 0.266 & 0.316 & 0.277 & 0.319 & 0.277 & 0.317 & 0.274 & 0.330 \\
                & 28 & 0.375 & 0.481 & 0.317 & 0.353 & 0.338 & 0.388 & 0.351 & 0.392 & 0.352 & 0.390 & 0.347 & 0.401 \\        
                
        \midrule  
        \multirow{4}{*}{\makecell[c]{TimeMixer\\ \cite{wang2024timemixer}}} & 7  & 0.181 & 0.309 & 0.077 & 0.124 & 0.095 & 0.150 & 0.077 & 0.129 & 0.076 & 0.133 & 0.102 & 0.169 \\
                & 14 & 0.249 & 0.358 & 0.125 & 0.168 & 0.160 & 0.200 & 0.109 & 0.160 & 0.110 & 0.163 & 0.151 & 0.210 \\
                & 21 & 0.449 & 0.523 & 0.182 & 0.222 & 0.240 & 0.288 & 0.176 & 0.216 & 0.170 & 0.214 & 0.244 & 0.293 \\
                & 28 & 0.512 & 0.610 & 0.218 & 0.265 & 0.302 & 0.351 & 0.206 & 0.249 & 0.191 & 0.238 & 0.286 & 0.343 \\
                
        \midrule
        \multirow{4}{*}{\makecell[c]{KAN\\ \cite{liu2024kan} }} & 7  & 0.176 & 0.268 & 0.085 & 0.131 & 0.101 & 0.146 & 0.076 & 0.130 & 0.071 & 0.123 & 0.102 & 0.160 \\
                & 14 & 0.257 & 0.374 & 0.126 & 0.169 & 0.148 & 0.203 & 0.116 & 0.168 & 0.110 & 0.165 & 0.151 & 0.216 \\
                & 21 & 0.292 & 0.441 & 0.144 & 0.196 & 0.160 & 0.225 & 0.138 & 0.200 & 0.133 & 0.199 & 0.174 & 0.252 \\
                & 28 & 0.357 & 0.502 & 0.165 & 0.216 & 0.188 & 0.258 & 0.185 & 0.243 & 0.178 & 0.240 & 0.214 & 0.292 \\
                
        \midrule
        \multirow{4}{*}{\makecell[c]{iTransformer\\ \cite{liu2023itransformer}}} & 7  & 0.205 & 0.320 & 0.080 & 0.120 & 0.083 & 0.135 & 0.072 & 0.118 & 0.070 & 0.111 & 0.102 & 0.161 \\
                & 14 & 0.282 & 0.416 & 0.121 & 0.165 & 0.131 & 0.184 & 0.113 & 0.161 & 0.105 & 0.150 & 0.150 & 0.215 \\
                & 21 & 0.310 & 0.455 & 0.127 & 0.175 & 0.154 & 0.212 & 0.129 & 0.178 & 0.123 & 0.171 & 0.169 & 0.238 \\
                & 28 & 0.335 & 0.503 & 0.156 & 0.206 & 0.193 & 0.263 & 0.160 & 0.215 & 0.144 & 0.201 & 0.198 & 0.278 \\

        \midrule
        \midrule
        \multirow{4}{*}{\makecell[c]{TimeGPT\\ \cite{garza2023timegpt}}} & 7  & 0.162 & 0.268 & 0.090 & 0.163 & 0.104 & 0.189 & 0.097 & 0.183 & 0.098 & 0.182 & 0.110 & 0.197 \\
                & 14 & 0.209 & 0.349 & 0.112 & 0.187 & 0.130 & 0.216 & 0.112 & 0.198 & 0.107 & 0.198 & 0.134 & 0.230 \\
                & 21 & 0.275 & 0.366 & 0.160 & 0.241 & 0.213 & 0.316 & 0.152 & 0.235 & 0.148 & 0.234 & 0.190 & 0.279 \\
                & 28 & 0.399 & 0.498 & 0.241 & 0.325 & 0.280 & 0.407 & 0.220 & 0.304 & 0.222 & 0.315 & 0.273 & 0.370 \\
                
        \midrule
        \multirow{4}{*}{\makecell[c]{TimesFM\\ \cite{das2023decoder}}} & 7  & 0.191 & 0.316 & 0.093 & 0.149 & 0.099 & 0.169 & 0.096 & 0.166 & 0.083 & 0.149 & 0.133 & 0.190 \\
                & 14 &  0.372 & 0.561 & 0.176 & 0.242 & 0.169 & 0.250 & 0.150 & 0.225 & 0.143 & 0.216 & 0.202 & 0.299 \\
                & 21 & 0.535 & 0.741 & 0.239 & 0.306 & 0.230 & 0.312 & 0.198 & 0.271 & 0.194 & 0.272 & 0.279 & 0.380 \\
                & 28 & 0.684 & 0.893 & 0.288 & 0.348 & 0.299 & 0.366 & 0.236 & 0.307 & 0.235 & 0.317 & 0.348 & 0.446 \\

        \midrule
        \multirow{4}{*}{\makecell[c]{Timer\\ \cite{liu2024timer}}} & 7  & 0.258 & 0.342 & 0.129 & 0.172 & 0.148 & 0.203 & 0.122 & 0.169 & 0.114 & 0.161 & 0.154 & 0.209 \\
                & 14 & 0.442 & 0.556 & 0.220 & 0.259 & 0.269 & 0.320 & 0.199 & 0.241 & 0.186 & 0.230 & 0.263 & 0.321 \\
                & 21 & 0.574 & 0.696 & 0.275 & 0.314 & 0.362 & 0.410 & 0.250 & 0.290 & 0.226 & 0.269 & 0.338 & 0.396 \\
                & 28 & 0.650 & 0.775 & 0.309 & 0.354 & 0.429 & 0.481 & 0.279 & 0.320 & 0.250 & 0.292 & 0.383 & 0.444 \\

        \midrule
                \multirow{4}{*}{\makecell[c]{Chronos  \\ \cite{ansari2024chronos}}} & 7  & 
                {0.125} & {0.224} & 
                {0.058} & {0.107} &
                {0.065} & {0.124} & 
                {0.065} & {0.124} & 
                {0.058} & {0.112} & 
                {0.074} & {0.138} \\
                & 14 & {0.198} & {0.343} & 
                {0.095} & {0.156} & 
                {0.103} & {0.179} & 
                {0.099} & {0.172} & 
                {0.086} & {0.158} & 
                {0.116} & {0.202} \\
                & 21 & {0.271} & {0.454} & 
                {0.133} & {0.204} & 
                {0.145} & {0.231} & 
                {0.135} & {0.217} & 
                {0.116} & {0.210} & 
                {0.160} & {0.261} \\
                & 28 & {0.337} & {0.536} & 
                {0.168} & {0.249} & 
                {0.186} & {0.280} & 
                {0.169} & {0.257} & 
                {0.141} & {0.232} & 
                {0.200} & {0.311} \\
        \midrule
        \midrule
        \multirow{4}{*}{\makecell[c]{\simraf \\ (Our method)} } & 7  
        & \textbf{0.116} & \textbf{0.208} & 
        \textbf{0.055} & \textbf{0.103} & 
        {0.066} & {0.125} & 
        \textbf{0.061} & \textbf{0.115} & 
        \textbf{0.057} & \textbf{0.107} & 
        \textbf{0.071} & \textbf{0.131} \\
                & 14 & {0.188} & {0.331} & 
                {0.088} & {0.148} & 
                {0.101} & {0.178} & 
                \textbf{0.091} & {0.160} & 
                {0.087} & {0.160} &
                {0.111} & {0.194} \\
                & 21 & {0.253} & {0.430} & 
                {0.117} & {0.185} & 
                {0.138} & {0.229} & 
                {0.122} & {0.198} & 
                {0.117} & {0.210} & 
                {0.150} & {0.247} \\
                & 28 & {0.320} &{0.512} & 
                {0.146} & {0.222} & 
                {0.173} & {0.276} & 
                {0.148} & {0.230} & 
                {0.142} & {0.224}&
                {0.186} & {0.293} \\

        \midrule
        \multirow{4}{*}{\makecell[c]{\miraf \\ (Our method) }} & 7  
        & {0.119} & {0.211} & 
        \textbf{0.055} & {0.103} & 
        {0.066} & {0.125} & 
        {0.062} & {0.117} & 
        {0.058} & {0.111} & 
        {0.072} & {0.133} \\
                & 14 & \textbf{0.181} & {0.325} & 
                \textbf{0.084} & \textbf{0.146} & 
                \textbf{0.098} & \textbf{0.177} & 
                \textbf{0.091} & {0.160} & 
                \textbf{0.084} & {0.156} &
            \textbf{0.107} & \textbf{0.193} \\
                & 21 & \textbf{0.241} & {0.428} & 
                \textbf{0.112} & {0.182} & 
                \textbf{0.130} & {0.224} & 
                \textbf{0.116} & {0.195} & 
                \textbf{0.109} & {0.192} & 
                \textbf{0.142} & {0.244} \\
                & 28 & \textbf{0.297} &{0.509} & 
                \textbf{0.136} & {0.215} & 
                \textbf{0.165} & {0.271} & 
                \textbf{0.138} & {0.225} & 
                \textbf{0.128} & {0.218}&
                \textbf{0.173} & {0.288} \\
        
        \bottomrule
    \end{tabular}
}
    
    \label{tab:main_result}
\end{table*}

\paragraph{Preprocessing.} Raw observations were obtained from operational monitoring systems and underwent standardized preprocessing before any forecasting or retrieval procedures were applied. Water-level measurements reported under different vertical datums: North American Vertical Datum of
1988 (NAVD88) and the National Geodetic Vertical Datum of 1929
(NGVD29), were converted to a common NGVD29 reference following the conversion factors available in EDEN. All variables were temporally aligned to ensure consistency across stations. The remaining missing values were handled using time-based interpolation and backward-fill strategies. 

\paragraph{Task Settings.} We ingest the lookback window of 100 days of all 38 variables as input, and forecast water levels at five gauging stations (NP205, P33, G620, NESRS1, NESRS2; see Figure~\ref{fig:everglade_map} in Appendix~\ref{sec:study_area}) over 7, 14, 21, and 28 day horizons. 
Retrieval-augmented forecasts use a pre-trained Chronos model without fine-tuning.

\subsection{Baselines}
We consider eight representative baseline methods across two types for the experimental comparison. 
\textbf{(1) Task-specific models}: DLinear \cite{zeng2023transformers}, TimeMixer \cite{wang2024timemixer}, Kolmogorov-Arnold Networks (KAN) \cite{liu2024kan} and iTransformer \cite{liu2023itransformer}.
\textbf{(2) Time series foundation models}: TimeGPT \cite{garza2023timegpt}, TimesFM \cite{das2023decoder}, Timer \cite{liu2024timer} and Chronos \cite{ansari2024chronos}.

\subsection{Evaluation Metrics}
Following the work \cite{shi2025deep}, we used Mean Absolute Error (MAE) (Eq. \ref{eq:mae}) and Root Mean Squared Error (RMSE) (Eq.~\ref{eq:rmse}) to evaluate the overall performance of these models. 
Additionally, we also include the Symmetric Extremal Dependence Index (SEDI) in (Eq. \ref{eq:sedi}) as a specialized metric for extreme values \cite{han2024weather,zheng2025sf}.
It classifies each prediction as either ``extreme'' or ``normal'' based on upper or lower quantile thresholds. 
SEDI values belong to the range $[0,1]$, where higher values indicate better accuracy in identifying extremes.
\begin{equation}
\text{MAE} = \frac{1}{N} \sum_{i=1}^N \left| y_i - \hat{y}_i \right|, 
\label{eq:mae}
\end{equation}
\begin{equation}
\text{RMSE} = \sqrt{\frac{1}{N} \sum_{i=1}^N \left( y_i - \hat{y}_i \right)^2}, 
\label{eq:rmse}
\end{equation}
\begin{equation}
\text{SEDI}(p) = \frac{\Sigma(\hat{y} < y_{\text{low}}^p \& y < y_{\text{low}}^p) + \Sigma(\hat{y} > y_{\text{up}}^p \& y > y_{\text{up}}^p)}{\Sigma(y < y_{\text{low}}^p) + \Sigma(y > y_{\text{up}}^p)},
\label{eq:sedi}
\end{equation}
where $y$ and $\hat{y}$ denote the ground truth and prediction, respectively. The tests, $\hat{y} > y_{\text{up}}^p$ and $y > y_{\text{up}}^p$, determine whether the predicted and observed values are extremes based on the threshold value $y_{\text{up}}^p$. Following the work \cite{han2024weather}, we set $10\%$ and $90\%$ as the low and high thresholds, respectively. $N$ is the number of samples.

\subsection{Implementation Details}
For task-specific models, we first train them before conducting the evaluation. 
The training is performed for a total of 1,000 epochs, starting with a learning rate of $1e^{-3}$. We used a batch size of 32 for training, and early stopping is executed if the training loss does not decrease for 50 iterations.
For pre-trained time series foundation models, we directly used their existing model weights for inference on our test set. 
For our retrieval-augmented forecasting approaches, a strong time series foundation, Chronos, is utilized as the forecaster.
All experiments were conducted on an NVIDIA A100 GPU with 80GB memory.
\section{Results and Analysis}

\subsection{Overall Comparison with Benchmarks}
Table~\ref{tab:main_result} reports the experimental comparison results between our retrieval-augmented forecasting methods, Sim-RAF and MI-RAF, and other representative baseline methods.
We highlight three interesting observations. 
First, Chronos exhibits superiority over other task-specific models and time series foundation models among baselines. Notably, both our Sim-RAF and MI-RAF approaches consistently and significantly outperform this strongest baseline. 
Second, for the temporal perspective, for short-term water level forecasting (lead time of 7 days), Sim-RAF yields an MAE and RMSE improvements by 4.1\% and 5.1\%, respectively, while MI-RAF boosts the performance by 2.7\% and 3.6\%, compared to Chronos.
Our methods show even greater improvements for long-term prediction: \textbf{Sim-RAF} presents MAE and RMSE improvements by 7.0\% and 5.8\%, respectively, while \textbf{MI-RAF} enhances the performance by 13.5\% and 7.4\%.
These results collectively underscore the power and potential of retrieval-augmented water level forecasting.
Last but not least, the performance of all models varies across the spatial locations, with the worst scenarios for the NP205 station. A possible reason for the lower performance at the NP205 station is its
weaker water-level correlation with other water stations (see Figure~\ref{fig:water_correlation} in Appendix~\ref{sec:correlation}). This suggests that dedicated domain knowledge is still needed to enhance the performance for particular stations. Inference runtime per test sample is approximately 0.04s for Chronos, 0.114s for Sim-RAF, and 0.162s for MI-RAF. Further experimental results can be found in Appendix \ref{sec:more_result}.

\subsection{Study on Extreme Events}
It is critical to study the model's performance on extreme cases.
Following the work \cite{han2024weather}, we use the SEDI metric to evaluate model performance under extreme events and report the results in Tables~\ref{tab:extreme_perform_7} and \ref{tab:extreme_perform_28}. The observations are as follows.
Compared to the strongest baseline,
\simraf and \miraf present an overall improvement by 4.4\% and 2.7\% for the predictions with a lead time of 7 days,
and exhibit an overall improvement by 14.1\% and 8.4\%, for the predictions with a lead time of 28 days.
These results show that both of our RAF approaches significantly outperform other baseline models, demonstrating the superiority of RAF-based methods for extreme values forecasting, in particular for longer lead time predictions. 
\begin{table}[ht!]
\centering
\caption{SEDI values (the higher, the better) for predicting extreme values across 5 stations with a lead time of 7 days. The best results are in \textbf{bold}.}
\resizebox{0.99\columnwidth}{!}{%
\begin{tabular}{l|ccccc|c}
\toprule
Models           & NP205 & P33 & G620 & NESRS1 & NESRS2 & Overall \\
\midrule
DLinear         & 0.475     & 0.456   & 0.456    & 0.523 & 0.478 & 0.477\\
TimeMixer       & 0.642     & 0.512   & 0.731    & 0.682 & 0.714 & 0.656\\
TimeGPT         & 0.300    & 0.217   & 0.152    & 0.095 & 0.086  & 0.170\\
TimesFM         & 0.550     & 0.043   & 0.065    & 0.119 & 0.086  & 0.172\\
Chronos         & 0.717     & 0.696   & \textbf{0.755}    & 0.651 & 0.690  & 0.702\\
\midrule
\simraf          & \textbf{0.760}     & \textbf{0.739}   & \textbf{0.755}    & \textbf{0.697} & \textbf{0.714}  & \textbf{0.733}\\
\miraf & 0.739     & 0.717   & \textbf{0.755}    & 0.674 & \textbf{0.714} & 0.720\\
\bottomrule
\end{tabular}%
}
\label{tab:extreme_perform_7}
\end{table}
\begin{table}[ht!]
\centering
\caption{SEDI values (the higher, the better) for predicting extreme values across 5 stations with a lead time of 28 days. The best results are in \textbf{bold}.}
\resizebox{0.99\columnwidth}{!}{%
\begin{tabular}{l|ccccc|c}
\toprule
Models           & NP205 & P33 & G620 & NESRS1 & NESRS2 & Overall \\
\midrule
DLinear         & 0.500     & 0.369   & 0.369    & 0.352 & 0.113 & 0.394\\
TimeMixer       & 0.119     & 0.049   & 0.341    & 0.000 & 0.000 & 0.102\\
TimeGPT         & 0.173     & 0.217   & 0.152    & 0.095 & 0.086  & 0.144\\
TimesFM         & 0.152     & 0.043   & 0.065    & 0.119 & 0.086  & 0.093\\
Chronos         & 0.619     & \textbf{0.738}   & 0.525    & 0.425 & 0.131  & 0.488\\
\midrule
\simraf         & \textbf{0.690}     & \textbf{0.738}   & \textbf{0.775}    & 0.425 & 0.158  & \textbf{0.557}\\
\miraf       & \textbf{0.690}     & 0.690   & 0.525    & \textbf{0.450} & \textbf{0.289} & \textbf{0.529} \\
\bottomrule
\end{tabular}%
}

\label{tab:extreme_perform_28}
\end{table}

\begin{figure}[ht!]
\centering
\includegraphics[width=0.99\columnwidth]{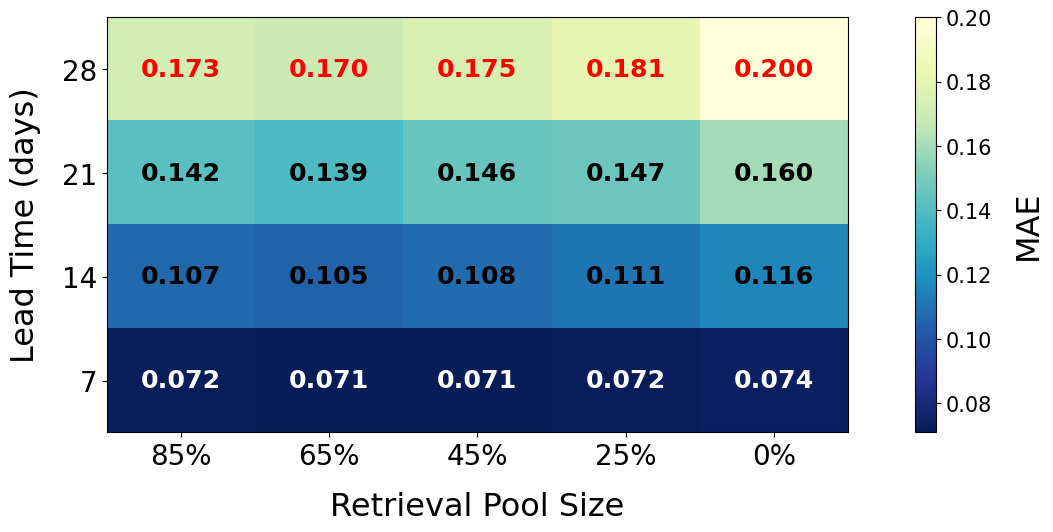}
\caption{Performance (MAE) vs.\ Retrieval Pool Size.}
\Description{Plot shows mean absolute error for different lead times (7, 14, 21, and 28 days) as a function of retrieval pool coverage from 0\% to 85\%.}
\label{fig:rafpool_study}
\end{figure}

\subsection{Effect of Retrieval Database Size}
To investigate the effect of the size of the retrieval pool, we compute mean absolute error (MAE) for five levels of retrieval pool sizes to cover $0\%$, $25\%$, $45\%$, $65\%$, and $85\%$ of the entire knowledge base across different lead times (7, 14, 21, and 28 days) in Figure~\ref{fig:rafpool_study}. 
We first observe that the $0\%$ retrieval coverage corresponds to the Chronos model without retrieval augmentation, which performs worst overall, especially at longer horizons. This again verifies the effectiveness and necessity of retrieval-augmented forecasting. 
More interestingly, we also noticed the MAEs decrease as the retrieval pool size increases when the coverage is less than $65\%$, which reveals that a larger retrieval pool can result in retrieving more contextual samples and benefit the subsequent water level forecasting tasks.
However, the performance drops slightly when the coverage reaches $85\%$.  
To better understand this behavior, we additionally examined retrieval similarity across different coverage levels. As the retrieval pool increased, the retrieved contexts generally became more similar to the query history, which helps explain why forecasting MAE initially decreases as coverage grows. However, this improvement does not continue in proportion at the highest coverage level. In this study, a larger retrieval pool refers to higher coverage of the historical knowledge base, and the 85\% setting represents the largest retrieval pool evaluated. At this level, forecasting MAE changes only marginally and slightly worsens compared with 65\% coverage. Future water levels in the Everglades are also influenced by rainfall variability, changes in gate and pump operations, and storage conditions. Therefore, expanding the retrieval pool too far can introduce contexts whose recent histories are statistically similar to the query, but whose subsequent future evolution may differ significantly due to operational or hydrological differences. A comprehensive justification for this observation is deferred to future work. Overall, these results empirically validate that retrieving relevant time series from a knowledge base significantly enhances predictive accuracy, particularly in long-range water level forecasting tasks. Additionally, the size of this retrieval base requires careful selection due to its substantial impact.

\subsection{Visualizations}
\begin{figure}[h]
\centering
    \includegraphics[width=0.87\columnwidth]{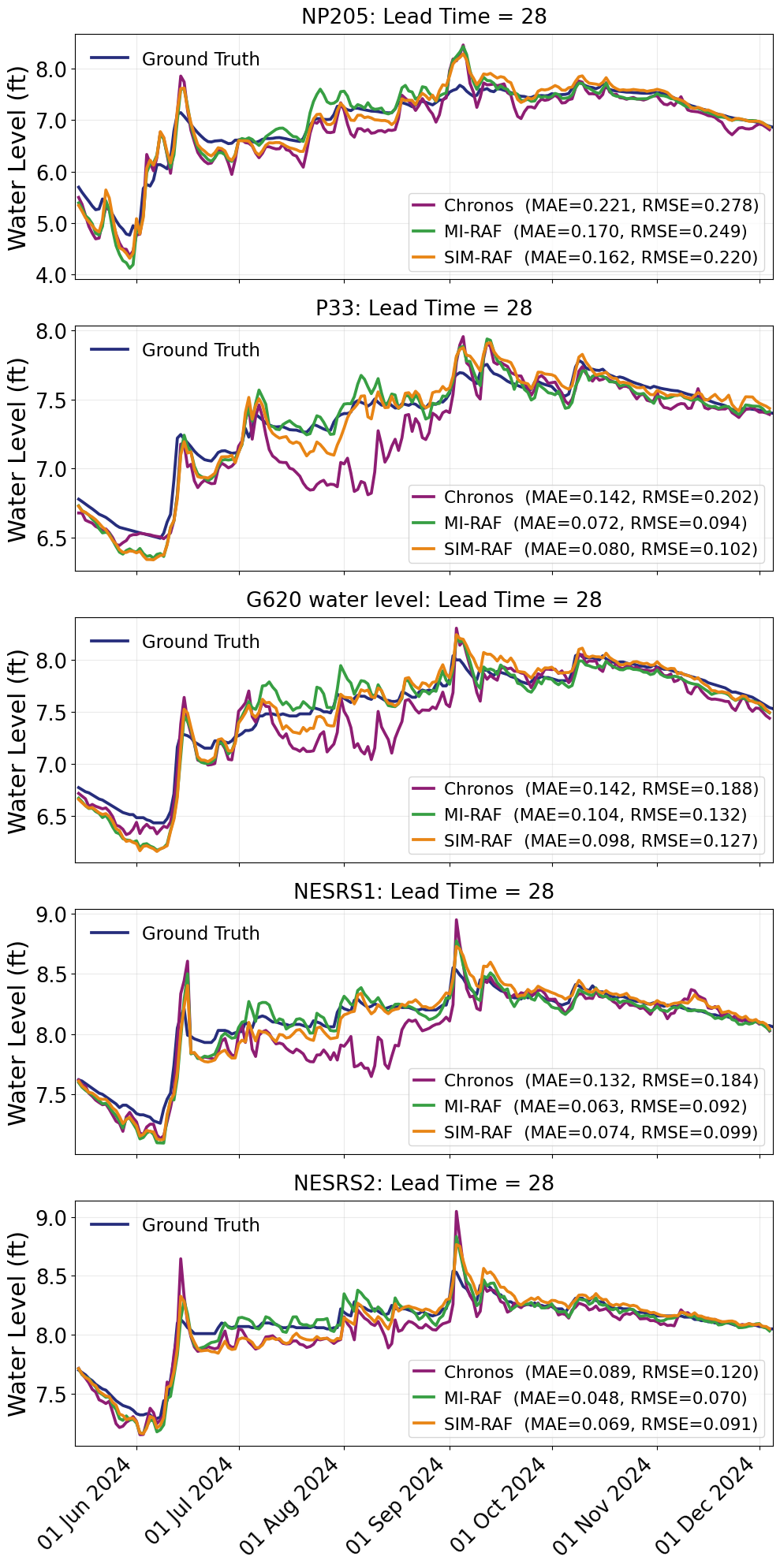}
\caption{Visualization of true and predicted water levels.}
\Description{Time series plot of observed water levels and three prediction curves (Chronos, Sim-RAF, and Mi-RAF) for a 28-day lead time. The Sim-RAF and Mi-RAF curves follow the observed series more closely than the Chronos baseline.}
\label{fig:raf28_P33}
\end{figure}

For the intuitive insights, we visualize the ground truth and predicted water levels in Figure~\ref{fig:raf28_P33}. 
We observed that the Chronos baseline exhibits noticeable deviations from the ground truth, particularly during periods of rapid fluctuation or extreme water levels. 
In contrast, both \simraf and \miraf show predictions that closely track the ground truth, demonstrating the superior accuracy over Chronos. 
This visual evidence supports our quantitative findings, illustrating how RAF approaches can better capture complex hydrological patterns and improve predictive accuracy.
Further visual results can be found in Appendix~\ref{sec:more_visual}.

\section{Discussion}
This study contributes an empirical characterization of how retrieval-augmented forecasting (RAF) changes the behavior of pretrained time-series foundation models in a real-world hydrological system. Relative to fixed-window forecasting, RAF improves performance most consistently at longer lead times (21 and 28 days) and under extreme water levels (defined as values outside the 10th and 90th percentile ranges).
The enhanced performance can be attributed to the highly relevant retrieved time series samples, which enrich the context available to the model and are crucial for the accurate prediction of these challenging long-term cases and rare events. 
The retrieval behavior also provides hydrologically meaningful insight into the Everglades system. The stronger gains observed at longer forecasting horizons suggest that water levels are influenced not only by short-term local dynamics but also by antecedent conditions, delayed basin-scale hydrological responses and operational controls. In contrast, the weaker improvements observed at NP205 are consistent with its lower correlation with other stations, as this site has different landscape characteristics and is profoundly influenced by operational deviations at certain times. This indicates that certain locations are influenced more strongly by landscape characteristics, operations of upstream structures, and hydrologic connectivity  than by shared basin-scale forcing.

These findings suggest that retrieval-augmented forecasting may benefit from more domain-aware and station-specific retrieval strategies.
For such hydrologically disconnected stations, constructing a knowledge base that incorporates a broader spectrum of relevant covariates may be more effective than relying solely on retrieval from historical time series data. 
Collaboration with domain experts is crucial for determining the relevant covariates. 
Furthermore, we examined the impact of varying the retrieval pool size on water level forecasting. Although larger retrieval pools generally improve performance, the optimal size requires careful selection.

\section{Conclusions}
To the best of our knowledge, this is the first systematic study of retrieval-augmented forecasting for water-level prediction in the Everglades, providing new insights into the behavior of foundation models under extreme hydrological conditions.
By leveraging historically relevant hydrological patterns, RAF consistently outperforms conventional forecasting approaches, particularly for long-horizon and extreme-event prediction.
Our findings demonstrate that retrieval-augmented approaches offer a scalable and computationally efficient pathway for water level forecasting in the Everglades.
The framework is transferable to other basins and hydrological systems with similar monitoring data, preserving the retrieval and evaluation protocol. More accurate forecasts directly support climate resilience, sustainable water resource management, and disaster mitigation efforts, aligning with several United Nations Sustainable Development Goals (e.g., SDG 6 “Clean Water and Sanitation,” SDG 13 “Climate Action,” and SDG 15 “Life on Land”). Future work can include hybrid retrievers, memory pruning for efficiency, and integration with physically informed models to further advance hydrological forecasting.

\section*{Interdisciplinary Collaboration and Impact}
ML and Everglades researchers were involved in multidisciplinary collaborations. Domain experts helped in area selection, data acquisition, data preprocessing, and analysis of the results. The proposed RAF methods, together with the accompanying web-based demonstration platform, can inform the development of decision-support tools used by government agencies, environmental managers, and emergency response teams. Accurate and interpretable water level forecasts can support flood preparedness, storm water release planning, drought mitigation, and the protection of ecologically critical wetlands.
Beyond the Everglades, the methodology supports broader societal impact by advancing environmentally sustainable forecasting systems and contributing to global efforts in climate adaptation and resilience. The combination of improved model performance, computational efficiency, and accessible visualization tools enables more transparent and responsible AI deployment in environmental science.

\section*{Limitations and Ethical Considerations} 
\textbf{Methodological Limitations.} RAF benefits from the presence of historically similar patterns. When such analogs are limited or absent, retrieval gains may diminish. 
\textbf{Data Privacy and Consent.}
EvergladesBench is built solely from publicly available operational monitoring sources (e.g., EDEN, DBHYDRO) collected by government agencies for environmental management and research. It contains no personally identifiable information and involves no human subjects; therefore, consent and privacy concerns do not apply.
\textbf{Bias and Representativeness.}
The dataset reflects the spatial coverage, sensor placement, and operational priorities of the Everglades monitoring network. Performance may vary across locations and hydrological regimes, and models may be biased toward conditions well represented in the historical record, while rare or unprecedented events remain challenging. 
\textbf{Responsible Use and Potential Misuse.}
The forecasting results produced in this study are intended for research and methodological evaluation. They should not be used as standalone decision-making tools in an operational setup. Any real-world application would need to be guided by domain experts and comply with relevant agency procedures.

\begin{acks}
This work is supported by the U.S. National Park Service under the award P24AC00474 and the U.S. National Science Foundation under the grant IIS-2331908. Any opinions, findings, conclusions, or recommendations expressed herein are those of the authors and do not necessarily reflect the views of these funding agencies.
\end{acks}

%% ------------------------------------------------------------------
%% Bibliography
%% ------------------------------------------------------------------
\bibliographystyle{ACM-Reference-Format}
\bibliography{sample-base} 

%% ------------------------------------------------------------------
%% Appendix
%% ------------------------------------------------------------------
% \newpage
\appendix
\section*{Appendix}

\label{sec:study_area}

\section{Study Area and Data Set}
\label{sec:dataset}
\subsection{Study Area}

\begin{figure}[ht]
\centering
\includegraphics[width=0.98\columnwidth]{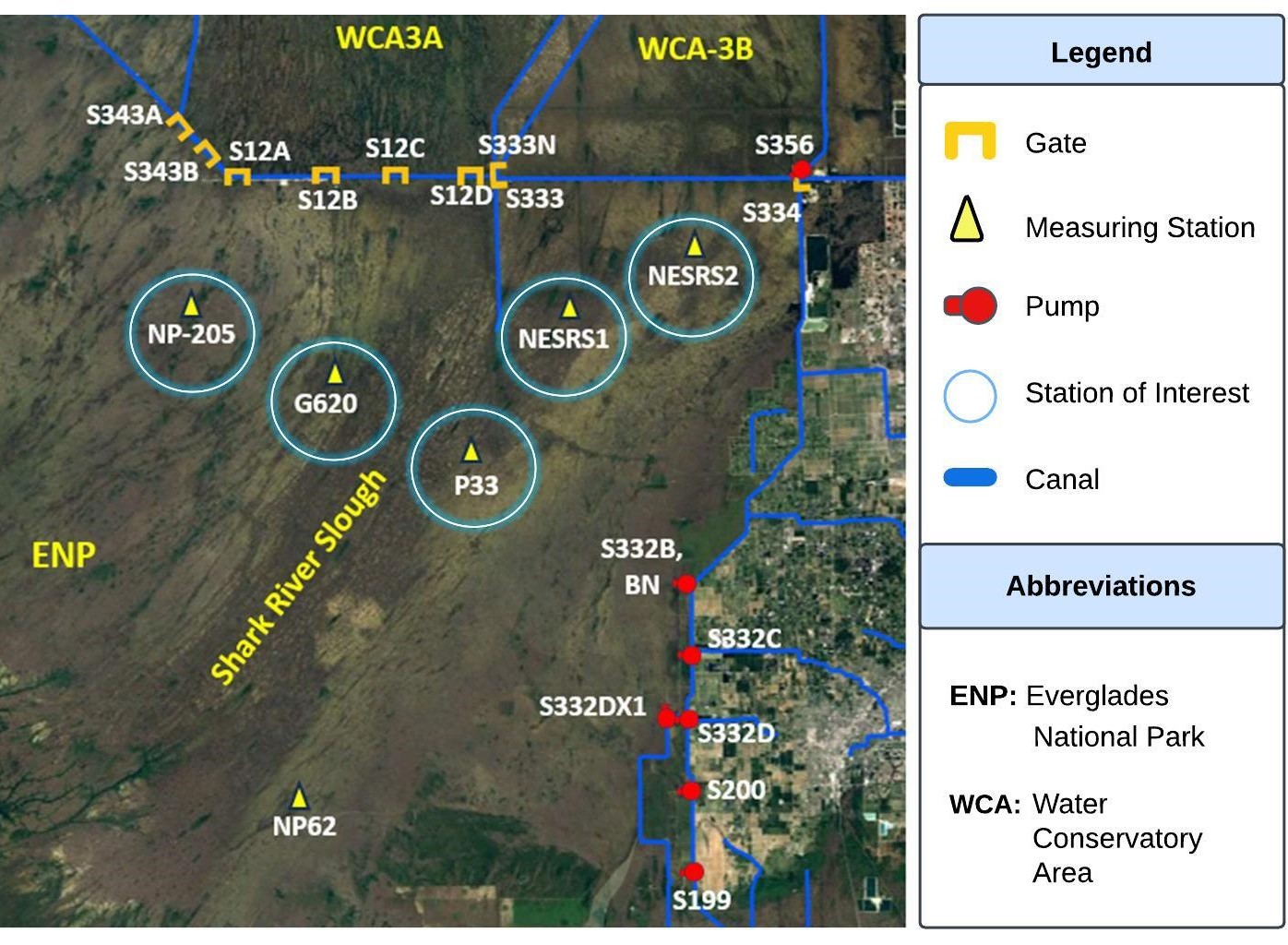}
\caption{Map of study area in Everglades National Park, highlighting the hydrological structures and monitoring stations.}
\Description{A map of Everglades National Park showing major canals, water control structures, and the five monitoring stations used in this paper. The locations are highlighted with markers to show their spatial distribution.}
\label{fig:everglade_map}
\end{figure}

\subsection{Data Summary}
Table~\ref{tab:data_summary} provides a summary of the key features included in the dataset used for the study. The data encompasses multiple hydrological and meteorological variables collected across various measuring stations. 
Daily rainfall data, measured in inches, is available for two measuring stations, namely NP205 and P33.
Potential Evapotranspiration (PET), measured in millimeters, represents the amount of water that would evaporate and transpire under normal conditions. PET is critical for understanding the water balance in the region as it helps to estimate the loss of water due to evaporation and transpiration. 
Daily pump and gate flow data is measured in cubic feet per second (cfs).  
Pumps and gates are used to regulate the water flow between different parts of the Everglades system, making the gate flow data essential for modeling water management operations. 
Daily water level data, recorded in feet is available for 14 stations.

\subsection{Correlation Study}
\label{sec:correlation}
To investigate the discrepancy between the NP205 station and other water stations, we compute their correlations. The right plot emphasizes correlations among NP205, P33, G620, NESRS1, and NESRS2, while the left plot extends this analysis to include nearby gates. The results reveal that NP205 exhibits lower correlations with the other stations, with values ranging from 0.76 to 0.85, whereas the remaining stations maintain higher correlations (at least 0.85).  
This aligns with the spatial distribution shown in Figure~\ref{fig:everglade_map}, where NP205 is positioned separately from the others. The hydrological flow dynamics, influenced by water management structures and natural flow patterns from WCA-3A and WCA-3B to Florida Bay, further explain these variations. 

\begin{figure}[ht!]
\centering
    \includegraphics[width=0.78\columnwidth]{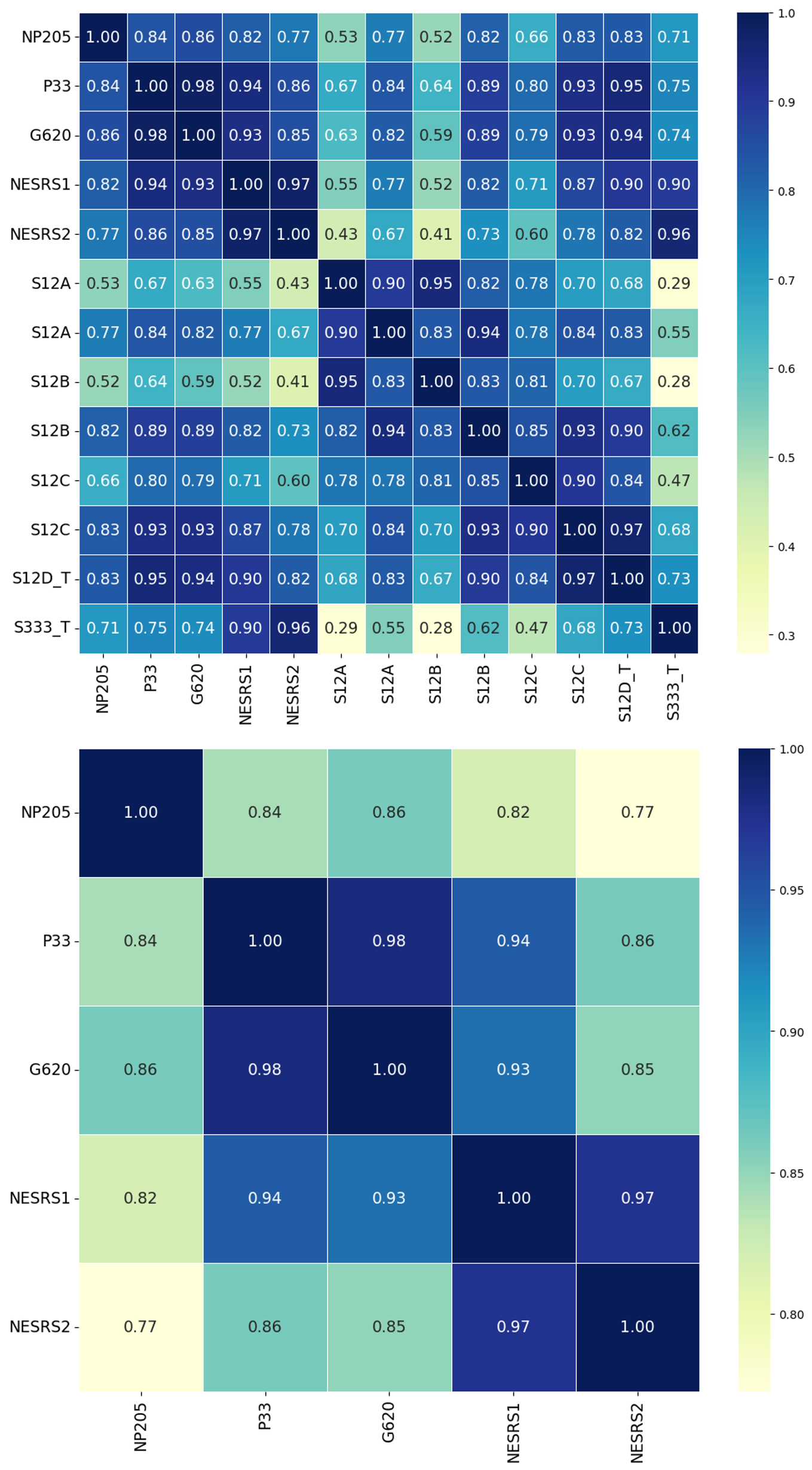}
\caption{Correlation Analysis of water levels at five measuring water stations and nearby gates.}
\Description{Two heatmaps showing Pearson correlations among hydrological structures. NP205 shows visibly lower correlation values compared to the other four stations, 
which appear in darker colors, indicating stronger correlations.}
\label{fig:water_correlation}
\end{figure}

\section{More Experimental Results}
\label{sec:more_result}

\begin{table}[H]
\centering
\caption{Stability of 7-day forecasting performance across five repeated runs. Values are reported as mean $\pm$ standard deviation across runs.}
\label{tab:stability}
\begin{tabular}{lcc}
\toprule
\textbf{Method} & \textbf{Avg. MAE} & \textbf{Avg. RMSE} \\
\midrule
Baseline & $0.0743 \pm 0.0000$ & $0.1384 \pm 0.0000$ \\
Sim-RAF  & $0.0714 \pm 0.0000$ & $0.1317 \pm 0.0000$ \\
MI-RAF   & $0.0724 \pm 0.0000$ & $0.1339 \pm 0.0001$ \\
\bottomrule
\end{tabular}
\end{table}

The repeated-run results show negligible variance across random seeds, indicating stable forecasting performance. A paired test on aligned forecast errors further shows that Sim-RAF improves over the baseline with statistical significance in the 7-day setting (paired $t$-test $p < 10^{-50}$; Wilcoxon signed-rank test $p = 2.3 \times 10^{-4}$).

\section{More Visualizations}
\label{sec:more_visual}
\begin{figure}[H]
\centering
\includegraphics[width=0.99\linewidth]{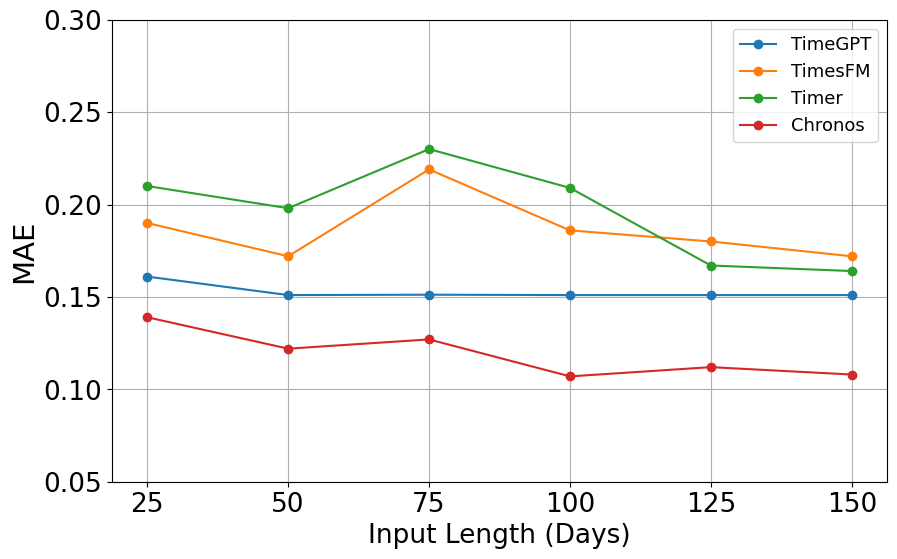}
\caption{Effect of input length on forecasting accuracy for time series foundation models 
Mean Absolute Error (MAE) is reported for a 28-day prediction horizon under varying input window lengths.}
\Description{...}
\label{fig:input_length_study}
\end{figure}

\begin{figure}[H]
\centering
\includegraphics[width=0.55\columnwidth]{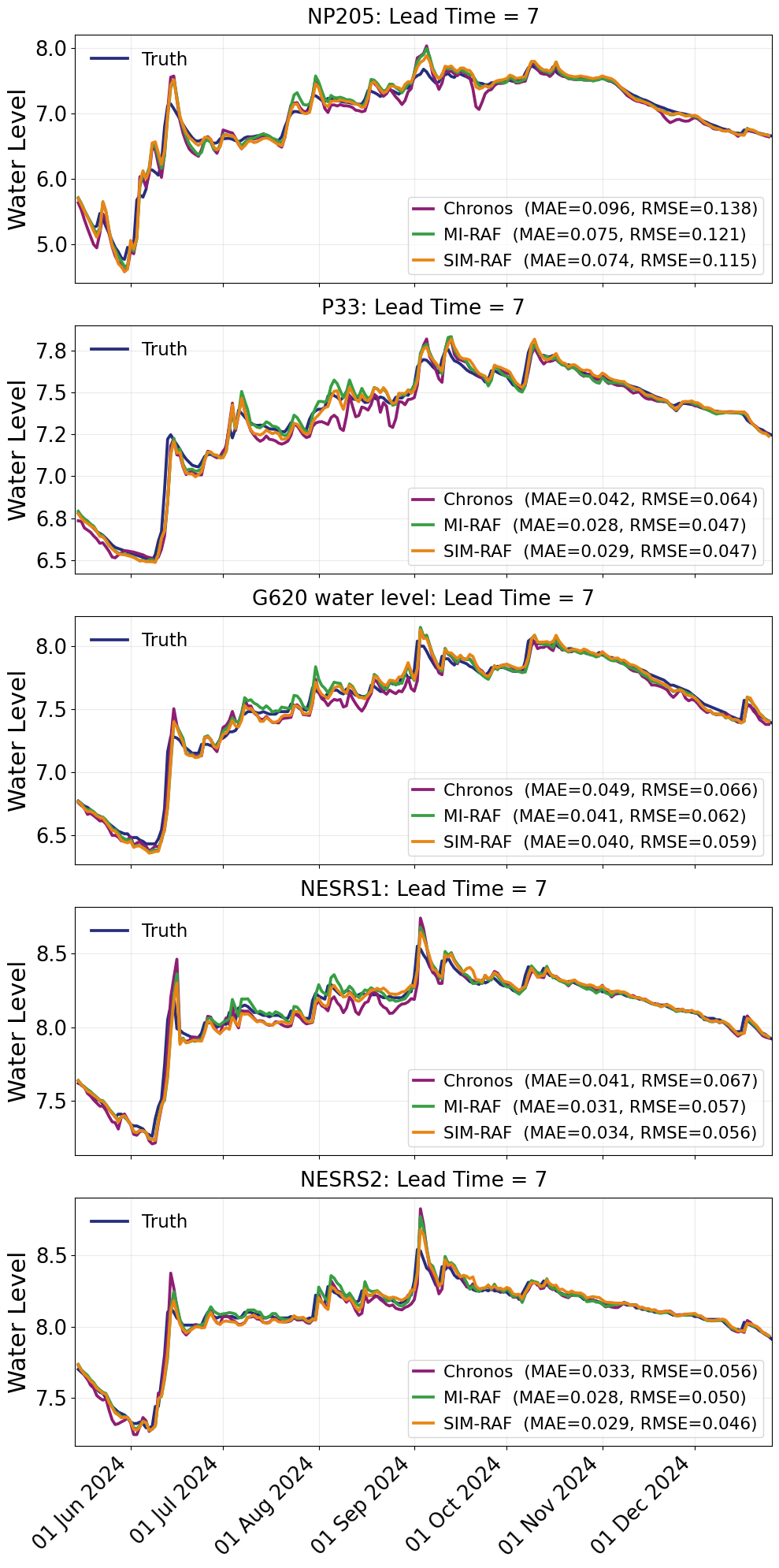}
\caption{Visualizations of true and predicted water levels for lead times of 7 days across 5 water stations, respectively.}
\Description{Line plots comparing observed water levels and the predictions made by Chronos, Sim-RAF, and MI-RAF for lead times of 7 days, respectively.}
\label{fig:raf_multi}
\end{figure}

\begin{figure}[H]
\centering
\includegraphics[width=0.55\columnwidth]{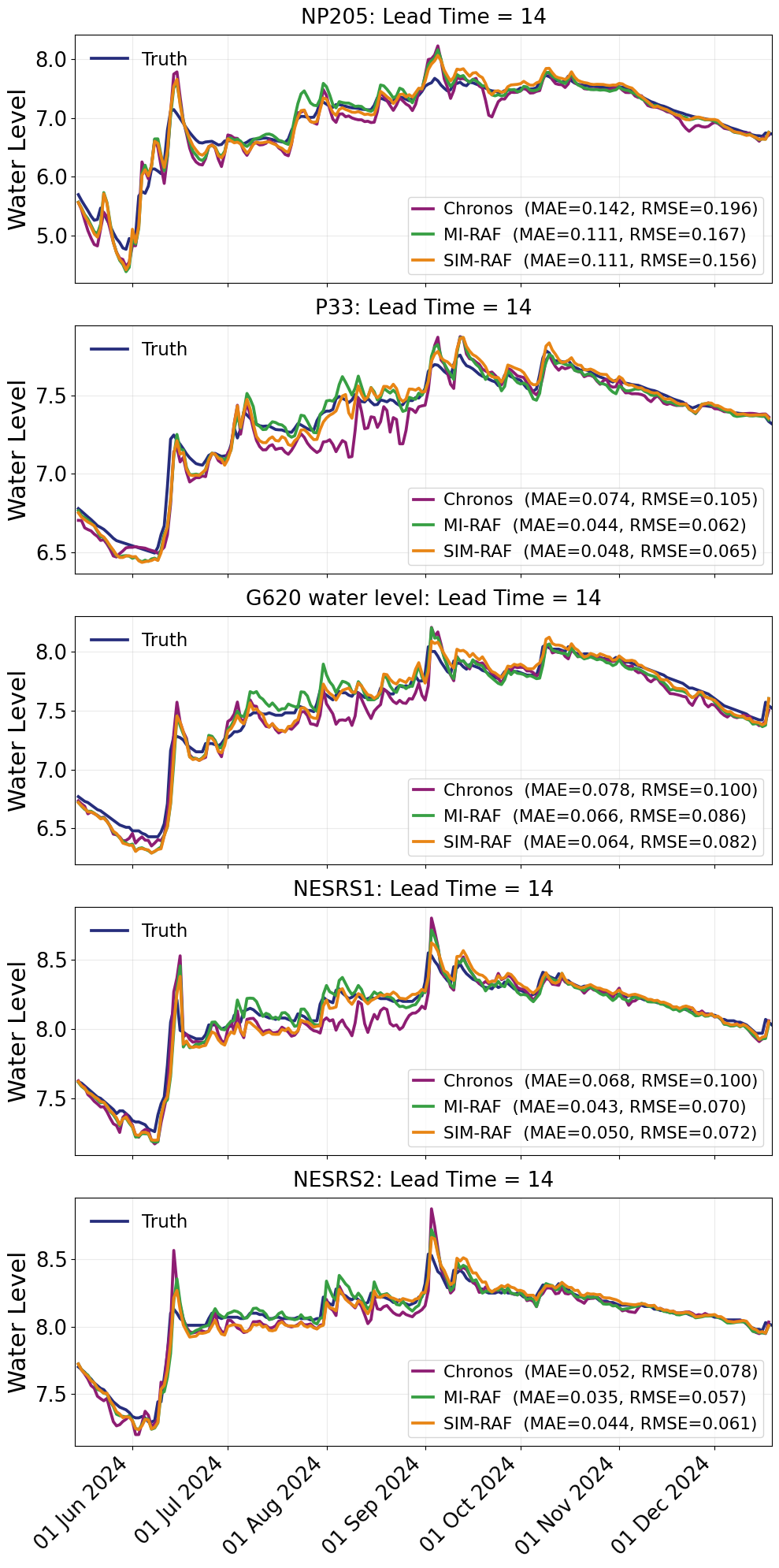}
\caption{Visualizations of true and predicted water levels for lead times of 14 days across 5 water stations, respectively.}
\Description{Line plots comparing observed water levels and the predictions made by Chronos, Sim-RAF, and MI-RAF for lead times of 14 days, respectively.}
\label{fig:raf_multi14}
\end{figure}

\begin{figure}[H]
\centering
\includegraphics[width=0.55\columnwidth]{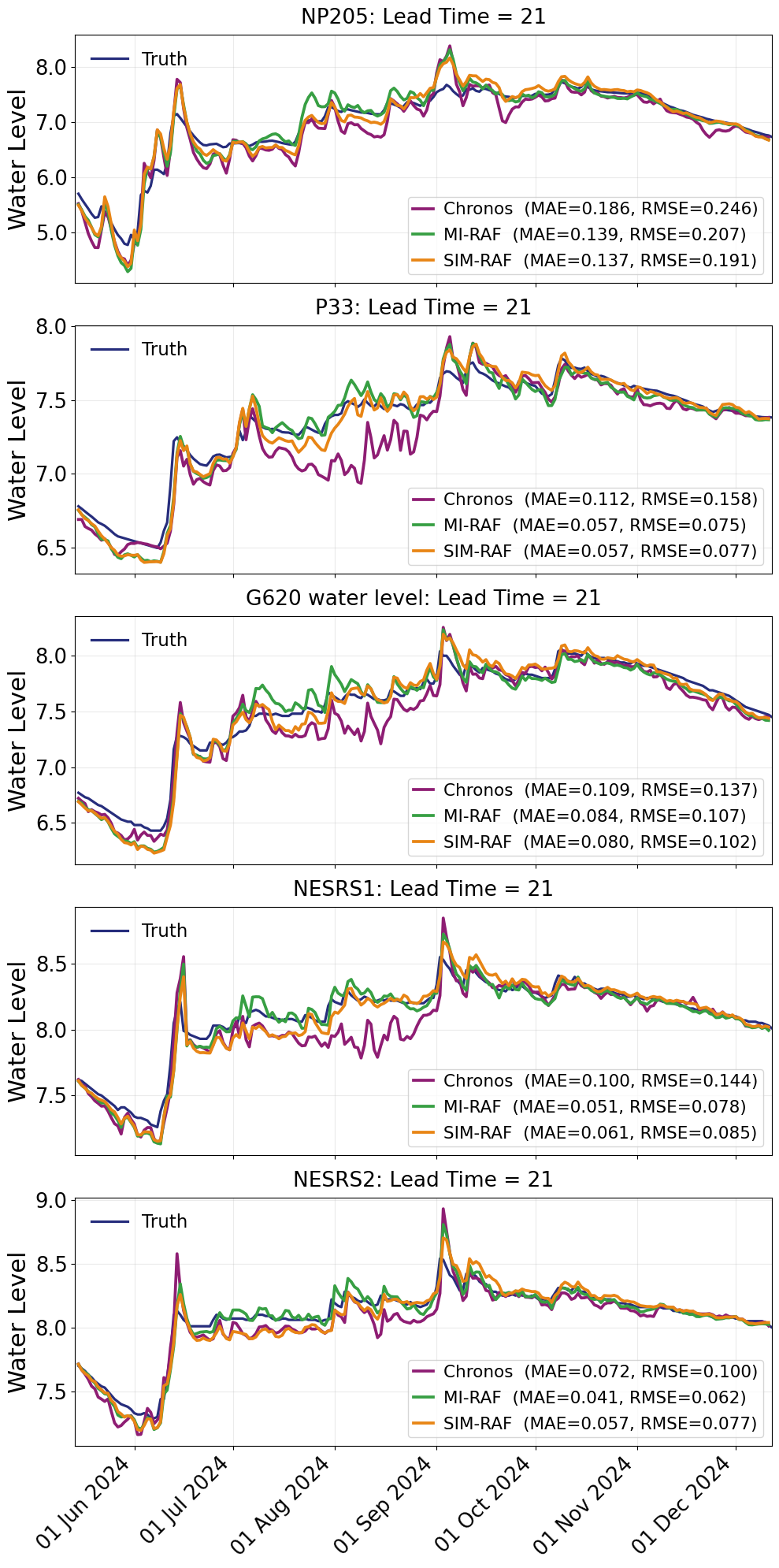}
\caption{Visualizations of true and predicted water levels for lead times of 21 days across 5 water stations, respectively.}
\Description{Line plots comparing observed water levels and the predictions made by Chronos, Sim-RAF, and MI-RAF for lead times of 21 days, respectively.}
\label{fig:raf_multi21}
\end{figure}

\begin{table*}[h]
\centering
\setlength{\tabcolsep}{2pt}        % tighter columns
\renewcommand{\arraystretch}{0.85} % tighter rows
\caption{Performance across 5 stations (NP205, P33, G620, NESRS1, NESRS2) for lead times of 7, 14, 21, and 28 days. The four models are our RAF methods with similarity-based and mutual information-based implementations utilizing the Strategy A and Strategy C techniques as mentioned in Section \ref{sec:augment}.}
    \resizebox{1.7\columnwidth}{!}{
    %\resizebox{0.8\textwidth}{!}{
    \begin{tabular}{c|c|cc|cc|cc|cc|cc|cc}
        \toprule
        \multirow{2}{*}{Models} & \multirow{2}{*}{\makecell[c]{Lead\\Time}} 
        & \multicolumn{2}{c|}{NP205} & \multicolumn{2}{c|}{P33} 
        & \multicolumn{2}{c|}{G620} & \multicolumn{2}{c|}{NESRS1} 
        & \multicolumn{2}{c|}{NESRS2} & \multicolumn{2}{c}{Overall} \\
        \cmidrule(lr){3-4} \cmidrule(lr){5-6} \cmidrule(lr){7-8} 
        \cmidrule(lr){9-10} \cmidrule(lr){11-12} \cmidrule(lr){13-14}
        & & MAE & RMSE & MAE & RMSE & MAE & RMSE & MAE & RMSE & MAE & RMSE & MAE & RMSE \\
                
        \midrule

        \midrule
        \multirow{4}{*}{\makecell[c]{\simraf \\ (Strategy A)} } & 7  
        &{0.117} & {0.206} & 
        {0.056} & {0.104} & 
        {0.069} & {0.127} & 
        {0.063} & {0.115} & 
        {0.057} & {0.107} & 
        {0.072} & {0.132} \\
                & 14 & {0.191} & {0.330} & 
                {0.090} & {0.151} & 
                {0.103} & {0.182} & 
                {0.094} & {0.161} & 
                {0.088} & {0.156} &
                {0.113} & {0.196} \\
                & 21 & {0.251} & {0.425} & 
                {0.118} & {0.189} & 
                {0.139} & {0.232} & 
                {0.126} & {0.201} & 
                {0.118} & {0.197} & 
                {0.151} & {0.249} \\
                & 28 & {0.320} &{0.519} & 
                {0.145} & {0.227} & 
                {0.175} & {0.283} & 
                {0.150} & {0.232} & 
                {0.139} & {0.224}&
                {0.186} & {0.297} \\

        \midrule
        \multirow{4}{*}{\makecell[c]{\miraf \\ (Strategy A) }} & 7  
        & {0.123} & {0.214} & 
        {0.057} & {0.106} & 
        {0.069} & {0.130} & 
        {0.063} & {0.119} & 
        {0.059} & {0.112} & 
        {0.074} & {0.136} \\
                & 14 & 
                {0.184} & {0.324} & 
                {0.087} & {0.151} & 
                {0.102} & {0.183} & 
                {0.092} & {0.162} & 
                {0.086} & {0.158} &
                {0.110} & {0.195} \\
                & 21 & {0.247} & {0.426} & 
                {0.115} & {0.187} & 
                {0.134} & {0.230} & 
                {0.120} & {0.200} & 
                {0.112} & {0.195} & 
                {0.146} & {0.248} \\
                & 28 & {0.304} &{0.519} & {0.141} & 
                {0.226} & {0.172} & 
                {0.281} & {0.142} & 
                {0.231} & {0.132}& {0.225}&
                {0.178} & {0.295} \\

        \midrule
        \midrule
        \multirow{4}{*}{\makecell[c]{\simraf \\ (Strategy C)} } & 7  
        & {0.126} & {0.221} & 
        {0.058} & {0.106} & 
        {0.068} & {0.126} & 
        {0.064} & {0.117} & 
        {0.059} & {0.109} & 
        {0.075} & {0.135} \\
                & 14 & {0.204} & {0.335} & 
                {0.093} & {0.152} & 
                {0.110} & {0.180} & 
                {0.098} & {0.164} & 
                {0.090} & {0.160} &
                {0.119} & {0.198} \\
                & 21 & {0.269} & {0.426} & 
                {0.118} & {0.179} & 
                {0.152} & {0.228} & 
                {0.124} & {0.190} & 
                {0.119} & {0.195} & 
                {0.157} & {0.246} \\
                & 28 & {0.330} &{0.500} & 
                {0.142} & {0.209} & 
                {0.175} & {0.265} & 
                {0.154} & {0.226} & 
                {0.149} & {0.228}&
                {0.190} & {0.290} \\

        \midrule
        \multirow{4}{*}{\makecell[c]{\miraf \\ (Strategy C) }} & 7  
        & {0.118} & {0.218} & 
        {0.054} & {0.104} & 
        {0.063} & {0.124} & 
        {0.059} & {0.115} & 
        {0.054} & {0.108} & 
        {0.070} & {0.133} \\
                & 14 & {0.184} & {0.326} & 
                {0.085} & {0.147} & 
                {0.097} & {0.175} & 
                {0.087} & {0.157} & 
                {0.082} & {0.156} &
            {0.107} & {0.193} \\
                & 21 & {0.237} & {0.408} & 
                {0.111} & {0.177} & 
                {0.128} & {0.214} & 
                {0.114} & {0.191} & 
                {0.106} & {0.188} & 
                {0.139} & {0.236} \\
                & 28 & {0.310} &{0.526} & 
                {0.137} & {0.214} & 
                {0.162} & {0.259} & 
                {0.137} & {0.223} & 
                {0.128} & {0.221}&
                {0.175} & {0.289} \\
                
        \bottomrule
    \end{tabular}
}
\label{tab:stratA_C_results}
\end{table*}

\end{document}